\documentclass[conference]{IEEEtran}
\IEEEoverridecommandlockouts
\usepackage{cite}
\usepackage{amsmath,amssymb,amsfonts}
\usepackage{graphicx}
\usepackage{textcomp}
\usepackage{xcolor}

\usepackage{algorithm}
\usepackage[noend]{algpseudocode}
\usepackage{setspace}
\usepackage{enumitem}

\def\BibTeX{{\rm B\kern-.05em{\sc i\kern-.025em b}\kern-.08em
    T\kern-.1667em\lower.7ex\hbox{E}\kern-.125emX}}
\begin{document}

\title{A Neural Rejection System Against Universal Adversarial Perturbations in Radio Signal Classification}

\author{Lu Zhang$^{1}$, Sangarapillai Lambotharan$^{1}$, Gan Zheng$^{1}$, Fabio Roli$^{2}$\\$^{1}$Wolfson School of Mechanical, Electrical and Manufacturing Engineering, Loughborough University,\\ Loughborough, UK\\$^{2}$Dept. of Electrical and Electronic Engineering, University of Cagliari, Piazza d’Armi, 09123 Cagliari, Italy}

\maketitle

\begin{abstract}
Advantages of deep learning over traditional methods have been demonstrated for radio signal classification in the recent years. However, various researchers have discovered that even a small but intentional feature perturbation known as adversarial examples can significantly deteriorate the performance of the deep learning based radio signal classification. Among various kinds of adversarial examples, universal adversarial perturbation has gained considerable attention due to its feature of being data independent, hence as a practical strategy to fool the radio signal classification with a high success rate. Therefore, in this paper, we investigate a defense system called neural rejection system to propose against universal adversarial perturbations, and evaluate its performance by generating  white-box universal adversarial perturbations. We show that the proposed neural rejection system is able to defend universal adversarial perturbations with significantly higher accuracy than the undefended deep neural network.
\end{abstract}


\section{INTRODUCTION}
Deep learning (DL) has gained huge success in a wide range of applications: image classification, object detection  \cite{redmon2017yolo9000, ren2015faster}, object recognition \cite{simonyan2014very, krizhevsky2012imagenet}, speech recognition \cite{saon2015ibm}, and language translation \cite{sutskever2014sequence}. Recently DL has also been successfully applied in automatic modulation classification (AMC) \cite{o2018over, scholl2019classification}. AMC plays an important role in signal intelligence and surveillance applications including that in cognitive radio and dynamic spectrum access to monitor the spectrum and its occupancy continuously. For many years, AMC has been implemented by extracting the signal features that are carefully crafted by the developers and by deriving compact decision boundary using low-dimensional extracted features. However, with the help of DL, AMC can be accomplished by directly training a deep neural network (DNN) with a large amount of raw signal data samples and making classification decisions with high confidence from the output of DNN. 

Despite the simplicity and good performance achieved by the DL in the AMC task, some researchers have discovered that a kind of new attack called adversarial examples can significantly deteriorate the classification performance of radio signals \cite{sadeghi2018adversarial}. Adversarial examples are generated by perturbing the original inputs with imperceptible and carefully designed modifications so as to mislead classification outputs of DNN. Precisely, adversarial examples can be formally described as follows. Given a trained DL classifier $f$ and an original input sample $x$, one can generate an adversarial example ${x}'$ as the solution to the constrained optimization problem below \cite{yuan2019adversarial}:

\begin{equation}
    \begin{aligned}
    \label{equ:adversarial examples}
    &~~~~\min_{x'}\left \| x'-x \right \|_{p}, \\
    &~~~~~s.t.~f\left ( x' \right ) = l',
    ~f\left ( x \right ) = l,
    ~l\neq l'.
    \end{aligned}
\end{equation}
where $l$ and $l'$ are the labels of $x$ and $x'$, respectively, and $\left \| \cdot  \right \|_{p}$ denotes the $l_{p}$-norm of the distance between two data samples. In the area of radio signal classification, the $l_{2}$-norm is appropriate as it represents the perturbation power \cite{sadeghi2018adversarial}.

\begin{figure}[ht]
\centering
\includegraphics[scale = 0.45]{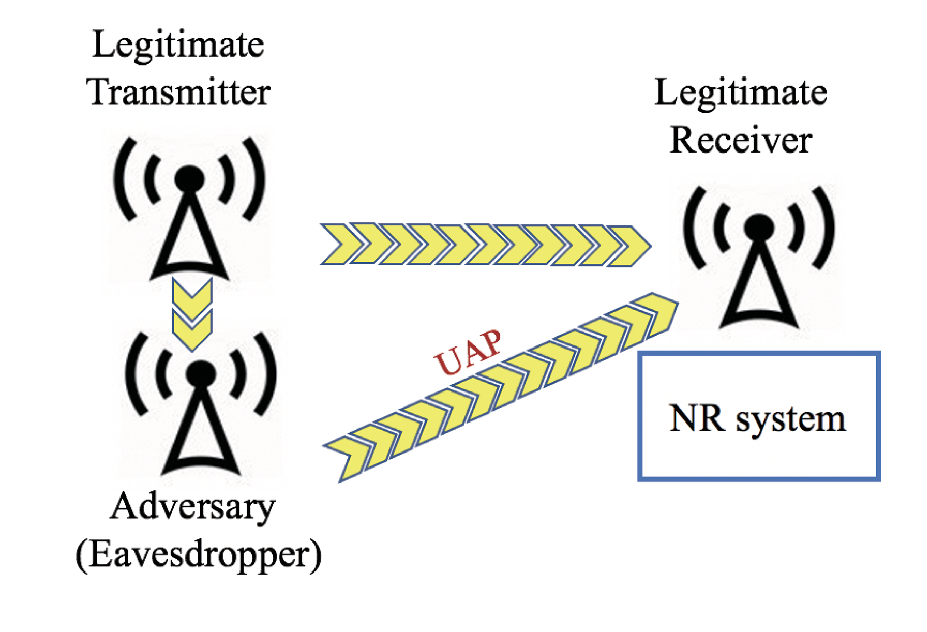}
  \caption{A civilian scenario for the adversarial examples in modulation classification.}
  \label{fig:eavesdropper}
\end{figure}

AMC has applications in both civilian and military domains. As illustrated in Figure \ref{fig:eavesdropper}, perturbation can be considered as a jamming strategy where an eavesdropper closer to the transmitter could decode the signal and transmit adversarial perturbations on the fly, assuming that the transmitter would continue with the same modulation type for a reasonable amount of time, i.e., spanning multiple symbols. At the legitimate receiver, the legitimate transmitted signals will be corrupted by the addition of perturbation signals which will lead to receiver misclassifications. However, this has a practical issue as the adversary needs to decode transmitted signals and generate perturbations instantly. This data dependent perturbation can be enhanced by using universal adversarial perturbation (UAP). In this way, perturbation is not instant data dependent and it is easier for the adversary to generate perturbations. 

The UAP attack was first proposed by Moosavi-Dezfooli et al. in \cite{moosavi2017universal}. In \cite{moosavi2017universal}, the authors showed that the existence of a universal (or image-agnostic) and very small perturbation vector $\eta$ that can cause natural images to be misclassified with a high probability $P$. The problem they formulated is to find a universal perturbation vector satisfying \cite{yuan2019adversarial}:
\begin{equation}
    \begin{aligned}
    \label{equ:UAP}
    &~~~~~~~~~\left \| \eta  \right \|_{p}\leq \varepsilon  \\
    &~P(f(x')\neq f(x))\geq 1-\delta 
    \end{aligned}
\end{equation}
where $\varepsilon$ limits the size of universal perturbation, and $\delta$ controls the
failure rate of all the adversarial samples. In \cite{sadeghi2018adversarial}, the authors showed that the UAP attack is able to largely decrease the classification accuracy of DNN in the AMC task. However, no defense was proposed against UAP attack for AMC. Therefore, in this paper, we propose to apply a neural rejection (NR) system \cite{sotgiu2020deep} against the UAP attack in the AMC problem. As illustrated in Figure \ref{fig:eavesdropper}, the NR system will be applied by the legitimate receiver to achieve better classification performance against the UAP attack. 

An NR system defends against adversarial examples through a rejection mechanism. The architecture of an NR system is composed of a pre-trained DNN and a support vector machine (SVM). Briefly, given an input signal $x$, the feature vectors are first extracted from the last feature layer of the DNN, then these vectors are fed into the connected SVM and the decision scores for this input signal $x$ are generated by the SVM classifier. Finally, if the maximum value of the generated decision scores is less than a pre-defined threshold, the input signal $x$ will be considered as an adversarial example and rejected. The main contribution of this work is that we propose a defense mechanism (NR system) in the area of radio signal classification and generate white-box UAP attack to evaluate the performance of the NR system. Through experiments, we show that the NR system is able to defend UAP attack with significantly higher classification accuracy than that of the undefended DNN.

The rest of the paper is structured as follows. Section \uppercase\expandafter{\romannumeral2} reviews the basics of adversarial examples. The proposed defense system is presented in Section \uppercase\expandafter{\romannumeral3} followed by the results and discussions in Section \uppercase\expandafter{\romannumeral4}. Finally conclusions are drawn in Section \uppercase\expandafter{\romannumeral5}.

\section{BASICS OF ADVERSARIAL EXAMPLES}
In this section, we introduce different approaches for generating adversarial examples, which can be categorized into three axes: threat model, perturbation, and benchmark \cite{yuan2019adversarial}. Here we mainly focus on the first two directions, i.e., threat model and perturbation. More details can be found in \cite{yuan2019adversarial}.

\subsection{Threat Model}
The attack model can define a number of potential attacks scenarios in terms of explicit assumptions on the adversarial falsification, adversary’s knowledge, adversarial specificity, and attack frequency. 

According to adversarial falsification, adversarial examples can be classified into false positive attacks and false negative attacks. False positive attacks generate a negative sample while being classified as a positive one. On the contrary, false negative attacks generate a positive sample while being classified as a negative one. The UAP attack adopted in this work belongs to the false negative attacks, where the benign signals perturbed by universal perturbations will be misclassified by the legitimate receiver as shown in Figure \ref{fig:eavesdropper}.

In terms of adversary's knowledge, adversarial examples can be categorized into white-box attack, black-box attack and grey-box attack. Specifically, white-box attack indicates the adversary is assumed to know everything about the targeted system, including the architecture and the parameters. Black-box attack indicates that the attacker does not have any knowledge about the defense system, while the grey box attack means that the attacker have part of the knowledge of the targeted system. In this work, all the attacks are white-box scenario as white-box attack is proved to be more effective than black-box and grey-box attacks.

As for adversarial specificity, adversarial examples can be classified into two different types: targeted attack and non-targeted attack. Targeted attack misleads the samples to a specified class, while non-targeted attack makes the sample fall into any wrong classes. This normally happens in a multi-class classification problem. For example, for a quadrature phase shift keying (QPSK) modulation signal, the targeted attack will force this QPSK modulation to be misclassified as a certain class, for instance binary phase-shift keying (BPSK), while the non-targeted attack will misclassify this QPSK modulation to be any other wrong classes. The UAP attack adopted in this work is non-targeted attack.

On the basis of attack frequency, adversary examples can be classified into one-time attack and iterative attack, where the former one takes only one step to optimize the attack and the latter one takes multiple times to optimize the attack. Normally the iterative attack can achieve a better fooling rate, which means that the iterative attack can be misclassified with higher success rate than the one-time attack. The UAP attack used in this paper belongs to the class of iterative attack.

\subsection{Perturbation}

Adversarial examples can also exploit different aspects of perturbation, including perturbation scope, perturbation limitation, and perturbation
measurement. 

According to perturbation scope, adversarial examples can be classified into individual attacks and universal attacks, where individual attacks generate adversarial perturbation for each data sample and universal attacks generate perturbation for the whole dataset. UAP attack used in this work is a typical universal attack. According to perturbation limitation, adversarial examples can be divided into optimized perturbation and constraint perturbation, where the former one sets perturbation as the goal
of the optimization problem and the other one sets perturbation as the constraints of the optimization problem. Furthermore, adversarial examples can also use different measurements to generate perturbations. In this context, $l_{p}$ norm is most commonly used in the literature. As mentioned in Section \uppercase\expandafter{\romannumeral1}, in the AMC problem, we choose the $l_{2}$ norm as it represents the perturbation power.

\section{THE PROPOSED NEURAL REJECTION SYSTEM AGAINST UNIVERSAL ADVERSARIAL PERTURBATION}
In this section, we introduce the defense mechanism used in this work, i.e., NR system, to defend against UAP attack. Moreover, the algorithm used to generate white-box UAP attack for NR system in AMC problem is also illustrated, which is to evaluate the performance of the proposed NR system.
\begin{figure}[ht]
\centering
\includegraphics[width = \columnwidth]{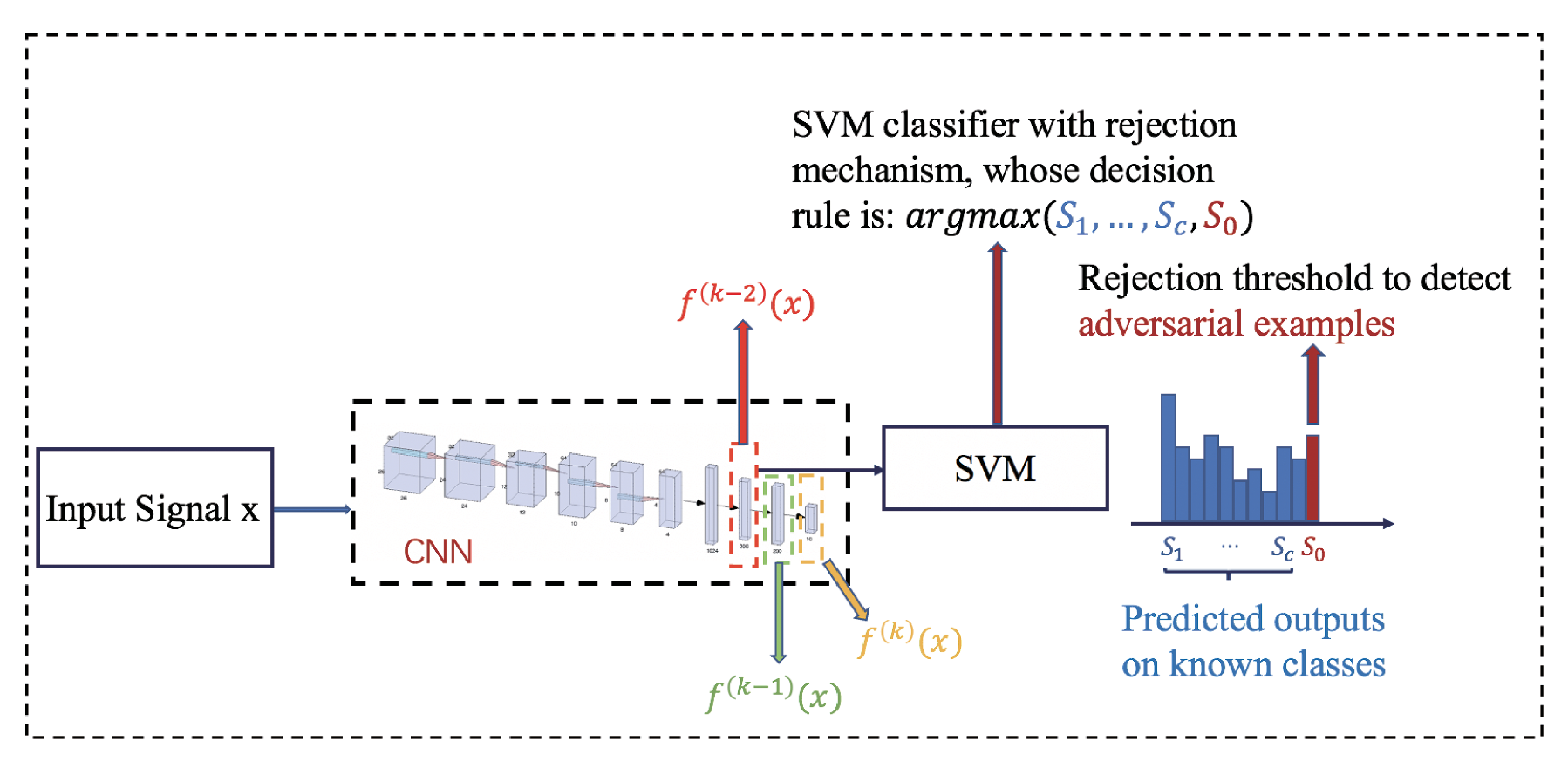}
  \caption{The architecture of the NR system. Given a data point $x$, the representations extracted from the last feature layer of CNN are fed into the connected one-vs-all SVM classifier. Then the decision scores of $x$, $(S_{1}, S_{2},...S_{c})$, will be generated from the output of the SVM classifier. If the maximum of the decision scores is less than the pre-defined threshold $S_{0}$, then the sample $x$ will be regarded as an adversarial example and be rejected.}
  \label{fig:NR}
\end{figure}

The architecture of the NR system shown in Figure \ref{fig:NR} is based on the work of \cite{sotgiu2020deep}. Before diving into the details, let us first introduce some notations of the the convolutional neural network (CNN). We define a CNN as a fuction $f: x\rightarrow y$. A CNN is composed of a set of perceptrons, and each perceptron maps a number of inputs to the output values with certain activation function (for example, ReLU function). The output of the CNN can be expressed as the following:
\begin{equation}
    \begin{aligned}
    \label{equ:cnn}
    f(x)=f^{(k)}(\cdot \cdot \cdot f^{(2)}(f^{(1)}(x)))
    \end{aligned}
\end{equation}
where $f^{(i)}$ is the function of the $i^{th}$ layer of the network, $i = 1, 2, \cdot \cdot \cdot, k$. Here, $f^{(k-1)}(x)$ is the logits layer of the CNN, and $f^{(k)}(x)$ represents a softmax layer which outputs the normalized decision scores, i.e., the sum of the outputs of the $f^{(k)}(x)$ layer will add up to 1. Furthermore, in the rest of the paper, we call the $f^{(k-2)}(x)$ layer as the last feature layer.

To train the NR system, we train the CNN as the first step, then the representations extracted from the last feature layer of CNN are used to train the connected SVM. The SVM classifier we use is a one-vs-all classifier which adopted the radial basis function (RBF) kernel.  Here, we introduce some notations for the SVM classifier. For an SVM, learning a linear classifier $S_{l}(x) = w^{T}x+b$ ($b$ is a bias term) is formulated as solving an optimization problem over the weights $w$:
\begin{equation}
    \begin{aligned}
    \label{equ:svm}
    \min_{w\in R^{d}}\left \| w \right \|^{2}+C\sum_{i}^{N}max(0,1-y_{i}S_{l}(x'_{i}))
    \end{aligned}
\end{equation}
where $C$ is a hyper-parameter in SVM to control the error, $y_{i}$ is the class label for each sample $x'_{i}$ and $N$ is the number of all training data samples. Then for an RBF-kernel SVM classifier, its decision scores (i.e., the output of the SVM) can be formulated as following:
\begin{equation}
\label{equ:decision function of SVM}
S(x)=\sum_{i=1}^{K}\alpha _{i}y_{i}exp(-\gamma \left \| x-x_{i} \right \|^{2})+b
\end{equation}
where $\alpha _{i}$ are dual variables, $\gamma$ is a RBF kernel parameter, $x_{i}$ is the support vector and $K$ is the total number of support vectors. Hence, to train a RBF-SVM classifier, we need to pre-set the hyper-parameters of SVM: $c$ and $\gamma$. In this work, grid search algorithm with three-fold cross validation is used to choose these two hyper-parameters. Here $c$ is chosen among $\left \{ 0.1,1.0,10.0 \right \}$ and $\gamma$ is chosen among $\left \{ 10^{-4}, 10^{-3},10^{-2} \right \}$. During the testing phase, given an input signal $x$, we take the features $\xi$ from the pre-trained CNN and these features are fed into the connected SVM. Through SVM, the decision scores of all the possible classes $S_{1}(\xi), S_{2}(\xi),...,{S_{c}(\xi)}$ for this input signal $x$ are generated as in (5). The decision function of the NR system is shown in (6), 
\begin{equation}
\label{equ:decision_function}
c^*=arg \max_{k=1,...,c}S_{k}(\xi), ~only ~if~S_{c^{*}}(\xi)>S_{0}
\end{equation}
which means the input signal $x$ will only be correctly classified when the maximum of the decision score $S_{c^{*}}$ is greater than a pre-defined threshold $S_{0}$. On the contrary, if the maximum of these decision scores are less than $S_{0}$, this input signal $x$ will be classified as an adversarial example and rejected.

Now we introduce the method to generate the white-box UAP attack for the NR system. The objective function to generate UAP attack for the NR system can be formulated as:
\begin{equation}
\label{equ:obejective function}
L(x,\cdot )=G_{y}(x)-G_{t}(x)
\end{equation}
where $G(x)$ is the decision score function of the NR system which is equal to $S(f^{(k-2)}(x))$, $y$ is the true class label and $t$ is the targeted class label. By minimizing $L(x,\cdot)$, the adversary is aiming to minimize the decision score for the true class while maximize the decision score for the targeted wrong class. Then the algorithm for generating UAP attacks for NR system is shown in Algorithm \ref{alg:UAP}, which is adopted from \cite{sadeghi2018adversarial}. Specifically, we first initialize the universal perturbation vector $v$ to a zero vector as in line 1. Then if the error rate of the perturbed data sets ($x_{v}$) is less than or equal to $1-\delta$, for each of the data point $x_{i}\in X$ which cannot evade the NR system (i.e., $S(x_{i}+v)=S(x_{i}) ~\mathbf{or} ~max(S(x_i+v))\leq S_{0}$), we compute the minimal perturbation $\Delta v_{i}$ using the fast gradient method (FGM) algorithm as shown in Algorithm \ref{alg:FGM} which sends $x_{i}+v$ to the decision boundary. Then we update the perturbation by projecting the updated perturbation $v+\Delta v_{i}$ into the $l_{p}$ norm ball of size $\varepsilon$. Here $\varepsilon$ is calculated as follows. 
\begin{equation}
    \begin{aligned}
    \label{equ:epsilon}
     \varepsilon =\sqrt{\frac{PNR}{SNR+1}}\cdot E(\left \| x \right \|_{2})
    \end{aligned}
\end{equation}
where PNR is the perturbation-to-noise ratio which indicates the ratio of the perturbation power to the noise power and SNR is the signal-to-noise ratio which means the ratio of the signal power to the noise power. Here, $E(\left \| x \right \|_{2})$ means the average of the $l_{2}$ norm of all the training samples. Finally, the algorithm stops until the error rate of the perturbed data sets ($x_{v}$) is greater than $1-\delta$.
\begin{algorithm}[ht]
\caption{Computation of Universal Adversarial Perturbations for NR System}\label{alg:UAP}
\hspace*{\algorithmicindent}\textbf{Input: }
\begin{itemize}[leftmargin=1.1cm]
\item data points $x$ and its true label $y$
\item the model $S(\cdot ,\theta )$
\item the rejection threshold $S_{0}$
\item desired $l_{p}$ norm of the perturbation $\varepsilon$
\item desired accuracy on perturbed samples $\delta$

\end{itemize}
\hspace*{\algorithmicindent}\textbf{Output: } Universal perturbation vector $v$.

\begin{algorithmic}[1]
\State Initialize $v\leftarrow 0$
\State \textbf{while} $Err(X_{v})\leq 1-\delta$  \textbf{do}
\State ~~~~~\textbf{for} each data point $x_{i}\in X$  \textbf{do}
\State ~~~~~~~~~~\textbf{if} $S(x_{i}+v)=y_{i} ~\mathbf{or} ~max(S(x_i+v))\leq S_{0}$ \textbf{then}
\State ~~~~~~~~~~~~~Compute the minimal perturbation that sends $x_{i}+v$ to the decision boundary:
\vspace{2mm}
\Statex  ~~~~~~~~~~~~~~~~~~~ $\Delta v_{i}\leftarrow arg \min_{r}\left \| r \right \|_{2} $
\Statex  ~~~~~s.t.~$S(x_{i}+v+r)\neq y_{i} ~\mathbf{and} ~max(S(x_i+v+r))>S_{0}$
\vspace{2mm}
\State ~~~~~~~~~~~~~~Update the perturbation:
\vspace{2mm}
\Statex  ~~~~~~~~~~~~~~~~~~~ $v\leftarrow P_{p,\varepsilon }(v+\Delta v_{i})$
\vspace{2mm}
\State ~~~~~~~~~~\textbf{end~if}
\State ~~~~~\textbf{end~for}
\State \textbf{end~while}
\end{algorithmic}
\end{algorithm}

Then we give details of line 5 in Algorithm \ref{alg:UAP}, which is to compute the minimal perturbation using the FGM algorithm for NR system as shown in Algorithm \ref{alg:FGM}. Given a data point $x$, for each possible class c, we first calculate the normalized perturbation direction $r_{norm}$ using line 4. The objective function $L(x,\cdot)$ is shown in (7). To calculate its gradient $\triangledown L(x,\cdot)$, we need to obtain the $\triangledown G(x,\cdot)$ by calculating the gradient of the last feature layer of CNN with respect to the input $x$, $\triangledown f^{(k-2)}(x)$, and the gradient of SVM $\triangledown S(\xi)$ respectively. To calculate $\triangledown f^{(k-2)}(x)$, the backpropagation algorithm \cite{rojas1996backpropagation} is adopted. Besides, we manually calculate $\triangledown S(\xi)$ as follows:
\begin{equation}
\label{equ:gradient of SVM}
\triangledown S(\xi)=\sum_{i=1}^{K}-2\gamma\alpha _{i}y_{i}exp(-\gamma \left \| \xi-\xi_{i} \right \|^{2})\cdot (\xi-\xi_{i})
\end{equation}
Finally, the chain rule is used to calculate the $\triangledown G(x,\cdot)$ as $\triangledown S(\xi) \cdot \triangledown f^{(k-2)}(x)$. After we obtain the normalized direction, the bisection search algorithm is adopted to find the minimum perturbation size using line 5 - line 13 in Algorithm \ref{alg:FGM}. Specifically, line 3 sets the maximum and minimum value of $\varepsilon $ as $p_{max}$ and $0$ respectively. Then for each time we calculate the average of the $\varepsilon $, $\varepsilon_{ave} $, as the middle point between $\varepsilon_{max} $ and $\varepsilon_{min} $ as in line 6. Then the data example will be updated by subtracting $\varepsilon_{ave}r_{norm}$. If the updated sample $x_{adv}$ cannot evade the NR system, i.e., the two conditions in line 8 are satisfied, then the minimum of $\varepsilon $ will be set as $\varepsilon_{ave} $ as in line 9. Otherwise, the maximum of $\varepsilon $ will be set to be $\varepsilon_{ave} $ as in line 11. The inner while loop will stop if $\varepsilon_{max} - \varepsilon_{min} $ is less than the desired perturbation accuracy. After iterating for each possible class c, we choose the target class as the class which has the minimum perturbation and the optimal perturbation size as the minimum of the perturbation. Finally, the adversarial perturbation of the input can be obtained using line 17.

\begin{algorithm}[ht!]
\caption{FGM-based Adversarial Perturbations for NR System}\label{alg:FGM}
\hspace*{\algorithmicindent}\textbf{Input: }
\begin{itemize}[leftmargin=1.1cm]
\item input $x$, true label $y$ and the number of classes C
\item the model $S(\cdot ,\theta )$ and the rejection threshold $S_{0}$
\item desired perturbation accuracy $\varepsilon _{acc}$
\item maximum allowed perturbation norm $p_{max}$

\end{itemize}
\hspace*{\algorithmicindent}\textbf{Output: }$\mathbf{r_{x}}$: adversarial perturbation of the input.

\begin{algorithmic}[1]
\vspace{2mm}
\State Initialize: $\varepsilon \leftarrow \mathbf{0}^{C\times 1}$

\State \textbf{for} $c$ in $range(C)$ \textbf{do}
\State ~~~$\varepsilon _{max}\leftarrow  p_{max},~ \varepsilon _{min}\leftarrow  0$
\State ~~~$r_{norm}=(\left \| \triangledown _{x}L(x, e_{c}) \right \|_{2})^{-1}\triangledown _{x}L(x, e_{c})$
\State ~~~\textbf{while} $\varepsilon _{max}- \varepsilon _{min}>\varepsilon _{acc}$ \textbf{do}
\State ~~~~~~$\varepsilon _{ave}\leftarrow (\varepsilon _{max}+\varepsilon _{min})/2$
\State ~~~~~~$x_{adv}\leftarrow x-\varepsilon _{ave}r_{norm}$
\State ~~~~~~\textbf{if} $S(x_{adv})==y ~\mathbf{or} ~max(S(x_{adv})) \leq S_{0}$  \textbf{then}
\State ~~~~~~~~~~~$\varepsilon _{min}\leftarrow \varepsilon _{ave}$
\State ~~~~~~\textbf{else}
\State ~~~~~~~~~~~$\varepsilon _{max}\leftarrow \varepsilon _{ave}$
\State ~~~~~~\textbf{end if}
\State ~~~\textbf{end while}
\State ~~~$[\varepsilon ]_{c} = \varepsilon _{max}$
\State \textbf{end for}
\State $t = arg \min_{\varepsilon ^{*}}\varepsilon $ and $\varepsilon ^{*} = min ~\varepsilon $
\State $r_{x}=-\frac{\varepsilon ^{*}}{\left \| \triangledown _{x}L(x, e_{t}) \right \|_{2}}\triangledown _{x}L(x, e_{t})$

\end{algorithmic}
\end{algorithm}

\section{RESULTS AND DISCUSSION}
\subsection{Experimental Setup}
\subsubsection{Dataset}
The dataset we used in this work is the GNU radio ML dataset RML2016.10a \cite{o2016radio}. The dataset contains 220000 input samples, and each sample corresponds to a specific modulation scheme. This dataset contains 11 different modulation schemes including BPSK, QPSK, 8PSK, QAM16, QAM64, CPFSK, GFSK, PAM4, WBFM, AM-SSB, and AM-DSB. These samples are crafted using 20 different SNR level from -20dB to 18dB with a step of 2dB. In this work, we use half of the dataset as the training set and the rest as the testing set. 

\subsubsection{Classifier}
In this work, we used the same CNN classifier as in \cite{sadeghi2018adversarial} which is called VT-CNN2 classifier. 


\subsubsection{Parameter Setting}
To generate UAP attacks for NR system, each time we randomly chose 50 samples that correspond to $SNR=10dB$ from the training set, and we test the accuracy of NR system against UAP attack using all the test samples which correspond to $SNR=10dB$ from the testing set, which consists of 5490 data samples. Among these 5490 samples, we call the samples that are correctly classified by the NR system as the set I samples. The experiments were repeated for 10 Monte Carlo trials and we calculated the average accuracy using these 10 Monte Carlo experiments. In every Monte Carlo experiment, we randomly chose 50 training samples to generate the UAP attack, therefore, the UAP attack generated for each time is a different vector. Then in each trial, we used the generated UAP vector to test the accuracy on the testing set. Finally we obtained an accuracy matrix which consists of the accuracy for 10 Monte Carlo trials, and we calculated the mean of the accuracy for these 10 trials. Furthermore, the threshold $S_{0}$ is chosen so that $10\%$ of the set I samples are rejected.

\subsection{Experimental Results}
\begin{figure}[ht]
\centering
\includegraphics[scale = 0.6]{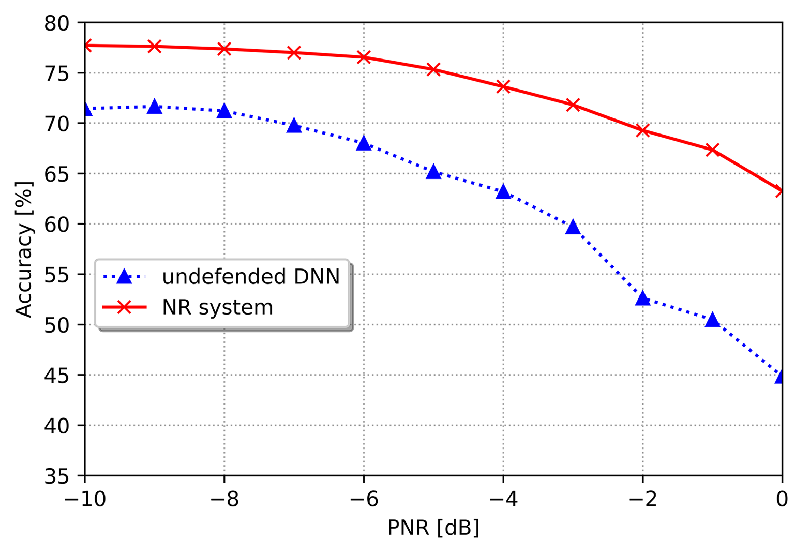}
  \caption{Accuracy of the NR system against white-box UAP attack as compared to that of the undefended DNN.}
  \label{fig:results}
\end{figure}

In this paper, we use undefended DNN against UAP attack as the benchmark. The accuracy of the NR system against white-box UAP attack as compared to that of the undefended DNN is shown in Figure \ref{fig:results}. Using the white-box scenario for both undefended DNN and the NR system, we can see that over a wide range of PNR level, NR system outperforms undefended DNN with a significant margin. And the improvement is more obvious when the PNR is large. For example, when PNR = 0dB, the accuracy of the NR system is almost 20$\%$ greater than that of the undefended DNN.

\section{CONCLUSIONS}
We have proposed a neural rejection based defense system to defend a modulation classification system against universal adversarial perturbations. By generating white-box universal adversarial perturbations for the neural rejection system, we have shown that the proposed neural rejection system can defend against universal adversarial perturbations with significantly higher accuracy than the undefended deep neural network. Approximately 20$\%$ better accuracy has been observed over a wide range of perturbation to noise ratio values.

\vspace{12pt}

\end{document}